\documentclass[a4paper]{article}

\usepackage{INTERSPEECH2022}

\usepackage{amssymb}
\usepackage{pifont}
\usepackage[dvipsnames]{xcolor}
\definecolor{mygreen}{RGB}{34,139,34}
\definecolor{myred}{RGB}{190,0,0}

\definecolor{clearlywhite}{RGB}{236, 249, 242}
\definecolor{almostwhite}{RGB}{216, 243, 229}
\definecolor{lightgreenaf}{RGB}{217, 242, 230}
\definecolor{lightergreen}{RGB}{179, 230, 204}
\definecolor{lightgreen}{RGB}{140, 217, 179}
\definecolor{mediumgreen}{RGB}{102, 204, 153}
\definecolor{lessdarkgreen}{RGB}{61, 194, 123}
\definecolor{darkgreen}{RGB}{57, 172, 115}
\definecolor{darkestgreen}{RGB}{49, 155, 98}

\newcommand{\cmark}{\ding{51}}
\newcommand{\xmark}{\ding{55}}
\usepackage{colortbl}

\title{DeToxy: A Large-Scale Multimodal Dataset for Toxicity Classification in Spoken Utterances}
\name{Sreyan Ghosh$^1$, Samden Lepcha$^3$, S Sakshi$^4$, Rajiv Ratn Shah$^2$, S. Umesh$^1$}

\address{
  $^1$Speech Lab, Department of Electrical Engineering, IIT Madras, Chennai, India\\
  $^2$MIDAS Labs, IIIT-Delhi, India \\
  $^3$TEG Analytics, Bangalore, India, $^4$Cisco Systems, Bangalore, India}
  
\email{gsreyan@gmail.com, sam.lepcha@outlook.com, ssakshi1397@gmail.com, rajivratn@iiitd.ac.in, umeshs@ee.iitm.ac.in}

\begin{document}

\maketitle
\begin{abstract}
Toxic speech, also known as hate speech, is regarded as one of the crucial issues plaguing online social media today. Most recent work on toxic speech detection
is constrained to the modality of text and written conversations with very limited work on toxicity detection from spoken utterances or using the modality of speech. In this paper,  we introduce a new dataset DeToxy, the first publicly available toxicity annotated dataset for the English language. DeToxy is sourced from various openly available speech databases and consists of over 2 million utterances. We believe that our dataset would act as a benchmark for the relatively new and un-explored Spoken Language Processing task of detecting toxicity from spoken utterances and boost further research in this space. Finally, we also provide strong unimodal baselines for our dataset and compare traditional two-step and E2E approaches. Our experiments show that in the case of spoken utterances, text-based approaches are largely dependent on gold human-annotated transcripts for their performance and also suffer from the problem of keyword bias. However, the presence of speech files in DeToxy helps facilitates the development of E2E speech models which alleviate both the above-stated problems by better capturing speech clues.


\end{abstract}

\section{Introduction}
\label{sec:intro}

Social network platforms are generally meant to share positive, constructive, and insightful content. However, in recent times, people often get exposed to objectionable content like threats, identity attacks, hate speech, insults, obscene texts, offensive remarks, or bullying. With the rise of different forms of content available online beyond just written text, i.e., audio and video, it is crucial that we device efficient content moderation systems for these forms of shared media. However, most prior work in literature and available datasets focus primarily on the modality of conversational text, with other modalities of conversation ignored at large. Thus, to alleviate this problem, in this paper, we propose a new dataset DeToxy, for the relatively new and unexplored Spoken Language Processing (SLP) task of toxicity classification in spoken utterances, which remains a crucial problem to solve for interactive intelligent systems, with broad applications in the field of content moderation in online audio/video content, gaming, customer service, etc. DeToxy is a large-scale multimodal dataset with both speech and text cues, manually annotated for toxicity detection in spoken utterances, with potential applications in building unimodal or multimodal models for conversational utterance-level Speech Toxicity Classification (STC) and End-to-End (E2E) speech systems which better capture semantics of the utterance.

The key challenge in toxicity classification in spoken speech is to learn good representations that capture toxicity signals from raw speech signals which is inherent in a rich set of acoustic and linguistic content, including semantic meaning, speaker characteristics, emotion, tone, and possibly even sentiment information while remaining invariant under different speakers, acoustic conditions, and other natural speech variations. However, different from Speech Emotion Recognition (SER) and similar to Speech Sentiment Recognition (SSER), toxicity classification from spoken utterances involve the system capturing semantic and syntactic properties of the utterance beyond just acoustic and prosodic cues which makes this task all the more difficult.

Traditional approaches for various speech downstream tasks involving spoken utterance classification employ low-level acoustic features, such as band-energies, filter banks, and MFCC features \cite{xie2019speech}, or raw waveform \cite{tzirakis2018end} . We acknowledge the fact that models trained on these low-level features can easily overfit to noise or signals irrelevant to the task. One way to remove variations in speech is to transcribe the audio into text and use text features to predict toxicity as done with SSER by \cite{lakomkin2019incorporating}. Nonetheless, toxicity signals in the speech, like tone and emotion, can be lost in the transcription. Thus, it is essential for any system to learn speech representations that make high-level information from speech signals available to solve downstream SLP tasks. Inspired by this, we provide both 2-step cascade and E2E baselines for STC on DeToxy and analyze the advantages and disadvantages in both, concluding why the modality of speech is crucial for the task of STC and makes DeToxy an important contribution to the speech community. Additionally, we show how prosodies and linguistic cues in natural speech aid our E2E model, which outperforms the 2-step methodology using less than 10\% of the total amount of labeled data available to our 2-step system.

\section{Related Work}
\label{sec:format}

Hate speech or toxic speech detection is a challenging task extensively studied in the literature, with techniques including  dictionary-based search, distributional semantics, and recent literature exploring the power of neural network architectures for the same. However, most of the work done on hate speech detection is constrained to just text, in English and some other foreign languages \cite{davidson2017automated,founta2018large,leite2020toxic} with the only existing work done by \cite{9616001} for audio.

On the other hand, the most commonly explored SLP tasks include Automatic Speech Recognition (ASR), SER , speaker verification, speaker identification, speech separation, speech enhancement, Named Entity Recognition (NER), phoneme recognition, etc, and the most recently explored speech sentiment classification which is very related to our task. All of these tasks are well studied with a lot of open-source datasets available online. With recent advancements in Natural Language Processing (NLP) and SLP, a lot of the systems achieving state-of-the-art in these tasks leverage E2E multi-layer neural networks like CNNs or Transformers, including self-supervised and semi-supervised techniques which have shown to help models in the low-resource data paradigm and proven to learn powerful speech representations \cite{baevski2020wav2vec,ling2020decoar,liu_2020}.


With downstream SLP tasks like speech sentiment classification and NER requiring an understanding of the contents and semantics of the spoken utterance, people have employed both 2-step \cite{cohn2019audio} and E2E methodologies \cite{yadav2020end,lu2020speech}. Some of these self-supervised methodologies, originally invented for pushing the performance of state-of-the-art in ASR systems, have also shown success in other downstream tasks \cite{liu_2020}.


\section{DeToxy and DeToxy-B}
\label{sec:dataset}

In this paper, we present DeToxy, the first annotated dataset for toxic speech detection in the English language. Our dataset is a subset of various open-source datasets, detailed statistics about which can be found in Table 1. We also present DeToxy-B, a balanced version of the dataset, curated from the original larger version taking into consideration auxiliary factors like trigger terms and utterance sentiment labels.

\subsection{Dataset Collection and Annotation}
For annotation, we define toxicity as \emph{rude, disrespectful, or otherwise likely to make someone leave a discussion}. For the initial version of our dataset, we primarily focus on openly available speech databases with or w/o text transcripts available. For obtaining transcripts of datasets for which transcripts were not available, we use pre-trained wav2vec-2.0 to obtain the transcriptions from a similar domain. In the process of consolidation of our dataset, primarily, after an empirical analysis of the transcripts of a majority of the open-source datasets available online, we found that most of them did not contain toxic utterances. Thus, we follow a 2-step approach to annotate our dataset. First, we train a textual toxicity classifier using \emph{BERT\textsubscript{BASE}} and use that to filter out datasets that had at least 10\% of their total number of utterances flagged as toxic by the model. Some datasets which did not fit this criterion were LibriSpeech \cite{panayotov2015librispeech}, TIMIT \cite{garofolo1993timit}, TED-LIUM \cite{hernandez2018ted} and MASS \cite{boito2019mass}. Next, all these utterances from all datasets filtered through step 1 were manually annotated by 3 professional annotators using audino \cite{DBLP:journals/corr/abs-2006-05236} taking both the textual transcript and speech files into account. Finally, we do a simple majority voting among the 3 annotations to determine the final toxicity class of each utterance. The Cohen's Kappa Score for inter-annotator agreement is 0.76. Table \ref{CS_tab} and Table \ref{CS_tab_2} show detailed descriptions of DeToxy and DeToxy-B datasets respectively.

\subsection{Sampling DeToxy-B and Test-Trigger Datasets}
For DeToxy-B, we keep all the toxic utterances from DeToxy and sample thrice the number of toxic utterances as our non-toxic utterances. We acknowledge the fact that sentiment is a strong controlling factor for various speech cues, and over-sampling any one of the sentiments might lead our E2E model to overfit the same. Thus, we don't sample at random but use sentiment labels to sample non-toxic utterances so that the final DeToxy-B dataset has an equal distribution of sentiments for its non-toxic utterances. These sentiment labels were annotated by a single annotator only for the purpose of constructing DeToxy-B and thus we don't include these labels in our final release of the dataset. In this paper, we evaluate all our proposed baselines on only DeToxy-B, with a Training, Development and Test stratified split of 70:15:15. 

Finally, we also collate an explicit test set with just non-toxic utterances where each utterance consisted of at least one trigger term \footnote{https://hatebase.org/}. We do this to evaluate how our baselines perform against not getting biased to trigger terms, a long-standing problem in toxicity classification \cite{sap-etal-2019-risk,garg2022handling}.

\begin{table*}[!ht]
\caption{DeToxy Statistics}
\label{CS_tab}
\begin{center}
\scalebox{1.0}{
\begin{tabular}{ c c c c c c c c c }
\hline
Dataset & \# Utterances  & \# Toxic & \# Non-Toxic & \# Sp & Toxic & Emo & Sent& TL (hh:mm:ss) \\ \hline
CMU-MOSEI \cite{zadeh2018multimodal} & 44,977 & 217  & 44,760 & 1,000  & \color{mygreen}\cmark &  \color{mygreen}\cmark & \color{mygreen}\cmark & 65:53:36 \\
\ CMU-MOSI \cite{zadeh2016mosi} & 2,199 & 67 & 2,132 & 98 & \color{mygreen}\cmark & \color{myred}\xmark &\color{mygreen}\cmark  &  02:36:17\\ 
\ Common Voice \cite{ardila2019common}& 1,584,219  & 2,888 & 1,581,331 & 66,173  &\color{mygreen}\cmark &\color{myred}\xmark   & \color{myred}\xmark & $\approx2181:00:00~~~$ \\ 
IEMOCAP \cite{busso2008iemocap} & 10,087  & 274   & 9,813 &  10 & \color{mygreen}\cmark & \color{mygreen}\cmark & \color{myred}\xmark & 11:28:12  \\ 
\ LJ Speech \cite{ljspeech17} & 13,100  & 40 & 13,060 & 1 & \color{mygreen}\cmark &\color{myred}\xmark & \color{myred}\xmark & 23:55:17   \\ 
\ MELD \cite{poria2018meld} & 13,708   & 142 & 13,566 & 304 & \color{mygreen}\cmark  & \color{mygreen}\cmark & \color{mygreen}\cmark & 12:02:44\\ 
\ MSP-Improv \cite{busso2016msp} & 8,348  & 129  & 8,219  & 13 &\color{mygreen}\cmark &\color{mygreen}\cmark  &\color{myred}\xmark  & 8:25:41 \\ 
\ MSP-Podcast \cite{lotfian2017building} & 73,042  & 692  & 72,350  & $\approx1,273~~~$ &\color{mygreen}\cmark &  \color{mygreen}\cmark & \color{myred}\xmark & 113:41:00\\ 
\ Social-IQ \cite{Zadeh_2019_CVPR} & 12,024  &  122 & 11,902 & - & \color{mygreen}\cmark &\color{myred}\xmark & \color{myred}\xmark & 20:39:00 \\ 
\ SwitchBoard \cite{godfrey1993switchboard} & 259,890&   456     & 259,434 & 400 & \color{mygreen}\cmark & \color{myred}\xmark & \color{myred}\xmark & $\approx260:00:00~~~$ \\ 
\ VCTK \cite{yamagishi2012english} & 44,583  & 50 & 44,533 & 110 & \color{mygreen}\cmark & \color{myred}\xmark & \color{myred}\xmark & 70:22:28 \\ \hline
\end{tabular}
}
\end{center}
\end{table*}

\begin{table*}[!ht]
\caption{DeToxy-B Statistics}
\label{CS_tab_2}
\begin{center}
\scalebox{1.0}{
\begin{tabular}{c c c c c c c c c  }
\hline
Dataset & \# Utterances  & \# Toxic & \# Non-Toxic & \# Sp &\# Positive & \# Neutral & \# Negative & TL (hh:mm:ss) \\ \hline
CMU-MOSEI  & 860  & 217  & 643 & - & 233 & 151 & 476 & 1:44:25 \\ 
CMU-MOSI  &  260 & 67  & 193 & - & 77 & 49 & 134 & 0:18:03 \\ 
Common Voice  & 11,551  & 2,888 & 8,663 & 6,350 & 3,241 & 2,199 & 6,111 & 12:38:17\\ 
IEMOCAP  & 1,090  & 274  & 816 & - & 313 & 195 & 582 & 1:19:26 \\ 
LJ Speech  & 148  & 40 & 108 & 1 & 30 & 27 & 91 & 0:14:57 \\ 
MELD  & 565  & 142  & 423 & $\approx67~~~$ & 145 & 136 & 284 & 0:31:05 \\ 
MSP-Improv & 523  & 129  & 394 & 12 & 115 & 91 & 317 & 0:36:32 \\ 
MSP-Podcast  & 2,772  & 692  & 2,080 &$\approx415~~~$ & 842 & 516 & 1,414 & 4:01:57\\
Social-IQ  & 479  & 122 & 357 & - & 147 & 93 & 239 & 0:36:40 \\ 
Switchboard  & 1,824  & 456  & 1,368 & - & 500 & 345 & 979 & 2:28:57\\ 
VCTK  & 199  & 50 & 149  & 81 & 58 & 33 & 108 & 0:08:51\\ \hline
\textbf{Total}  & \textbf{20,271}  & \textbf{5,077} & \textbf{15,194} & \textbf{-} & \textbf{5,701} & \textbf{3,835} & \textbf{10,735} & \textbf{24:39:10} \\ \hline

\end{tabular}}
\end{center}
\end{table*}

\section{Baselines}
\label{sec:baselines}

\subsection{Problem Formulation}

The task of toxicity detection from spoken utterances involves assigning a probability score, denoting the toxicity, to each utterance fed to the model. Formally put, if \textbf{X} = \{$x_1$, $\cdots$, $x_i$, $\cdots$, $x_N$\} are the set of input utterances from the dataset $D$ with $N$ utterances, then \textbf{Y} = \{$y_1$, $\cdots$, $y_i$, $\cdots$, $y_N$\} are their corresponding labels where \emph{y\textsubscript{i}} $\in \{0,1\}$.

In the next 2 sections we describe in detail our proposed 2-step and End-to-End baselines for DeToxy, the different components involved, and the training procedure involved in each.

\subsection{2-step Baselines}
\label{sssec:twostep}

Our 2-step approach consists of 2 primary components, an ASR component that produces transcriptions from spoken utterances, and a Sequence Classification Model for classifying the toxicity of the output transcript. For our ASR model, we resort to using wav2vec-2.0 \cite{baevski2020wav2vec}. Wav2vec-2.0 is built on the transformer architecture and has a Convolutional Neural Network (CNN) feature encoder layer, which encodes raw audio into speech frames. Wav2vec-2.0 follows the pre-training and fine-tuning paradigm wherein we first pre-train the model using a large unlabeled audio corpus in a self-supervised fashion solving a contrastive task as follows:

\begin{equation}
\label{eqn:wav2vec_1}
\mathcal{L}_{m}=-\log \frac{\exp \left(\operatorname{sim}\left(\mathbf{c}_{t}, \mathbf{q}_{t}\right) / \kappa\right)}{\sum_{\tilde{\mathbf{q}} \sim \mathbf{Q}_{t}} \exp \left(\operatorname{sim}\left(\mathbf{c}_{t}, \tilde{\mathbf{q}}\right) / \kappa\right)}
\end{equation}

where $\mathcal{L}_{m}$ is the InfoNCE loss that wav2vec-2.0 tries to minimize, $\mathbf{c}_{t}$ is the context representation of a masked frame obtained from the last transformer layer, $\mathbf{q}_{t}$ is the quantized representation of obtained through a product quantization based quantization module and $\tilde{\mathbf{q}}$. Precisely, for every masked input frame which is input from the feature encoder module to the transformer-based context encoder, wav2vec-2.0 tries to identify the true quantized representation among a set of distractors. For more implementation-specific details about wav2vec-2.0, we refer our readers to \cite{baevski2020wav2vec}. For our 2-step approach, we use the wav2vec-2.0\textsubscript{LARGE} architecture to push performance on ASR.

Post SSL pre-training, the model is then fine-tuned by minimizing the CTC objective function on labeled speech data. We do not use a Language Model in our final setup to decode our utterances during inference, as we did not find significant differences in Word Error Rate (WER) with it in our experiments, and also to keep the setup simple.

Both self-supervised and supervised training paradigms for ASR tend to get biased to the domain from where the data originates \cite{hsu2021robust} and therefore we make careful data considerations for pre-training and fine-tuning our model. Most works in literature report results on a single benchmark dataset and so we resort to multiple dataset pre-training but single dataset fine-tuning, We choose to pre-train our model using a combination of Libri-Light, Common Voice (CV) \cite{ardila2019common}, SwitchBoard (SWBD) \cite{godfrey1993switchboard}, and Fisher \cite{cieri2004fisher}. These 4 datasets cover a diverse set of domains and suit well to our downstream ASR fine-tuning needs without creating a huge domain-shift \cite{hsu2021robust}. For downstream ASR fine-tuning, we resort to 3 different dataset setups as follows: 1) To fit our model to the conversational domain, we fine-tune it on 300 hours of SWBD . 2) For read speech we resort to 960hrs of LibriSpeech fine-tuning and finally 3) We also fine-tune on CV which comprises of a majority of DeToxy.


For our text-based sequence-classification model, which is the second step of our 2-step approach, we employ the \emph{BERT\textsubscript{BASE}} architecture from the transformer family. Since its inception, transformers have achieved SOTA on sequence classification tasks including toxicity or hate-speech classification \cite{malik2021toxic,zhao2021comparative}. Formally put, we tokenize each word in the sentence and feed it as input through the transformer architecture. Next, we utilize the hidden state embedding \emph{e} corresponding to the \emph{[CLS]} token, where \emph{e} $\in \mathbb{R}^{768}$, as the aggregate representation of words in the transcript of the utterance. This embedding  \emph{e} is now fed through a final classification layer which learns a parameterized function $f(.)$ and outputs $f(e) = h_x \in \mathbb{R}^{n}$ where $n=2$. Finally, we pass the logits obtained through a softmax activation function to get the probability distribution $p$ of toxicity for each sentence. For training our model we minimize the Cross Entropy (C.E.) loss as follows:


\begin{equation}
\label{eqn:cross_entropy}
\text {C.E.}=-\frac{1}{N} \sum_{i=1}^{N} \sum_{j=1}^{M} y_{i_j} \log \left(p_{i_j}\right)
\end{equation}

where $y_{i_j}$ is corresponds to the $j^{th}$ gold annotated class from the $i^{th}$ training instance in our dataset and $p_{i_j}$ is the probability score given by the model that the $i^{th}$ instance belongs to the $j^{th}$ class. During inference, we simply perform the $argmax(p_i)$ to find the final class for each utterance.

We fine-tune \emph{BERT\textsubscript{BASE}} on a publicly available large-scale hate-speech classification dataset \cite{do2019jigsaw} or on the relatively smaller gold toxicity annotations from the DeToxy-B dataset.

\subsection{End-to-End Baselines}
\label{sssec:e2e}

For our End-to-End (E2E) Baselines, we choose to experiment with different feature extractors, where we build baselines employing either low-level feature extractors (eg. Filter-Banks) or high-level features from contextual feature encoder models learned through SSL on unlabelled speech data. In the next 2 sub-sections, we briefly describe both our approaches.

\subsubsection{F-Bank}
\label{sssec:fbank}

For our Filter-Bank (F-Bank) based E2E approach, given an audio input sequence $x$, we extract log-compressed mel-scaled F-Banks with a window size of 25 ms and a hop size of 10 ms to obtain \emph{T} frames. Post this, we employ $GlobalAveragePooling(.)$ over all the \emph{T} frames [\emph{h\textsubscript{1}}, $\cdots$,\emph{h\textsubscript{T}}], where $h_t$ corresponds to the frame at time-step $t$, to obtain a single embedding \emph{e} $\in \mathbb{R}^{80}$ for each utterance. This embedding is then fed into a toxicity classification decoder, consisting of a single fully connected layer post which we employ the \emph{softmax} activation function to the output of this layer to obtain the probability of toxicity for each utterance.

\begin{figure}[t]
\centering
\includegraphics[width=0.4\textwidth]{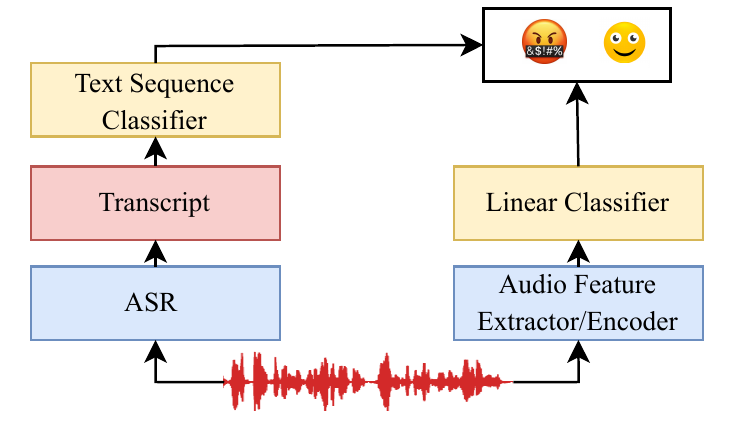}
\caption{Proposed Baselines: 2-step approach (left) and E2E approach (right)}
\label{Sequence Tagger Model}
\end{figure}

\subsubsection{wav2vec-2.0}
\label{sssec:subsubhead}

For a fair comparison with our 2-step approach, we employ the same wav2vec-2.0 architecture employed in the 2-step procedure for our E2E system and experiment with only the \emph{BASE} architecture. We follow the same pre-training methodology as our 2-step baseline ASR system, but this time we fine-tune the parameters of our model on DeToxy-B, solving a downstream toxicity classification task, and train in an E2E fashion. During downstream training, we either freeze or keep all the parameters of wav2vec-2.0 trainable. In both cases, the CNN feature extractor remains frozen by default. 

To fine-tune our model on the downstream task, we employ the same pooling strategy and \emph{prediction-head} as our F-Bank approach, except now we take the embeddings from the last layer of our ASR encoder and obtain a single embedding \emph{e} $\in \mathbb{R}^{768}$ for each utterance by passing it through a $GlobalAveragePooling(.)$ layer which pools over all the time-steps.

Both these models are trained using Cross-Entropy Loss (\ref{eqn:cross_entropy}) and similar to our 2-step procedure, during inference, we take the $argmax(.)$ of the output of the $softmax(.)$ function as the final class for each utterance.

\section{Experimental Setup}
\label{sec:exp_setup}
We use the PyTorch  Framework for building and evaluating all our Deep Learning models along with the Transformer implementations, pre-trained models, and specific tokenizers are taken from the HuggingFace library. We train all our models with Adam optimizer in batched mode with a batch size of 8 and a learning rate of $1e^{-4}$  for a maximum of 100 epochs with an early-stopping of 5 epochs. We make all our code and data publicly available on GitHub\footnote{https://github.com/Sreyan88/Toxicity-Detection-in-Spoken-Utterances}

\section{Results and Result Analysis}
\label{sec:typestyle}

Table \ref{tab:my_label}, provides the \emph{macro-averaged F\textsubscript{1}} scores obtained for variations of both our \textbf{2-step} and \textbf{E2E} baselines on the Development, Test and Test-T splits of DeToxy-B post training on the Train split. The category section of 2-step system is denoted as ``Transcript Source - Sequence Classifier Data Source'', where Gold is gold transcripts and LS, CV, and SWBD stand for fine-tuning on LibriSpeech, Common Voice, and SwitchBoard respectively. CC \cite{do2019jigsaw} and DB stand for \emph{BERT\textsubscript{BASE}} fine-tuned on Civil Comments or DeToxy-B gold text transcripts respectively. Beyond freezed and un-freezed wav2vec-2.0 E2E setups, based on findings by \cite{shah2021all}, we also test our E2E model by taking embeddings from the $9^{th}$ layer which contains maximum semantic information.



\begin{table}[h!]
    \caption{Experimental Results of our 2-step and E2E baselines on DeToxy-B}
    \label{tab:my_label}
    \centering
    \begin{tabular}{l l l l l}
    \hline
    System & Category & Dev & Test & Test-T \\
    \hline
    2-step  & Gold-CC & \cellcolor{mediumgreen}0.787 & \cellcolor{mediumgreen}0.801 &  \cellcolor{clearlywhite}0.258\\
    & Gold-DB & \cellcolor{darkestgreen}0.921 & \cellcolor{darkestgreen}0.934 & \cellcolor{lightgreen}0.492\\
    & LS-CC &   \cellcolor{lightgreen}0.672 & \cellcolor{lightgreen}0.682 &   \cellcolor{lightgreenaf}0.281\\
    & LS-DB & \cellcolor{mediumgreen}0.742 & \cellcolor{mediumgreen}0.744 &\cellcolor{lightergreen}0.484 \\
    & CV-CC & \cellcolor{lightergreen}0.659 & \cellcolor{lightgreen}0.674 & \cellcolor{almostwhite}0.270 \\
    & CV-DB & \cellcolor{lightgreenaf}0.505 &\cellcolor{mediumgreen}0.724 & \cellcolor{lightergreen}0.474 \\
    & SWBD-CC & \cellcolor{lightergreen}0.672 &\cellcolor{lightgreen}0.682 & \cellcolor{almostwhite}0.270 \\
    & SWBD-DB & \cellcolor{mediumgreen}0.726 &\cellcolor{mediumgreen}0.737 & \cellcolor{lightergreen}0.478 \\ \hline
    F-Bank  & -   &  \cellcolor{lightergreen}0.610 &\cellcolor{lightgreen}0.620 & \cellcolor{lightgreen}0.491  \\  
    wav2vec-2.0  & Freezed    & \cellcolor{almostwhite}0.448  &\cellcolor{lightgreenaf}0.457 & \cellcolor{mediumgreen}0.497  \\
    & Un-freezed  & \cellcolor{lessdarkgreen}0.877 & \cellcolor{lessdarkgreen}0.869  & \cellcolor{lessdarkgreen}0.510 \\
    wav2vec-2.0 (9) & Un-freezed  & \cellcolor{darkgreen}0.897 & \cellcolor{darkgreen}0.877 &\cellcolor{darkestgreen}0.540 \\\hline
    \end{tabular}
\end{table}

As we see in Table 3, our E2E system clearly outperforms our \textbf{2-step} system setup when gold transcripts are not available to the text sequence classifier. We hypothesize the true reason to be that ASR models don't generalize across domains and generally fail to perform well when trained on one domain and inferred on another. This is also very evident in our case where we get an average WER of 33\%, 43\% and 26.9\% on LS, CV and SWBD respectively. \textbf{2-step} systems trained on CC consistently under-performs, which reveals that text sequence classifiers models trained to classify toxicity are not robust to changes in domain. 

Our un-frozen wav2vec-2.0 E2E setup with representations taken from the $9^{th}$ layer outperforms all our true baselines. This proves that semantic information of the utterance is crucial to toxicity classification from spoken speech. This setup also outperforms other baselines on Test-T, in which very evidently all our baselines struggled on, especially our 2-step baselines trained on CC.




\section{Conclusion}
\label{ssec:subhead}
In this paper, we introduce the first publicly available human-annotated dataset DeToxy for the task of toxic speech classification and propose two strong baselines for this task. Future work includes expanding the dataset at least 10 fold using more naturally spoken utterances, exploring different modalities, and developing better neural architectures to solve this problem.

\bibliographystyle{IEEEtran}

\bibliography{template}

\end{document}